\documentclass[sigconf]{acmart}
\pdfoutput=1
\AtBeginDocument{%
  \providecommand\BibTeX{{%
    \normalfont B\kern-0.5em{\scshape i\kern-0.25em b}\kern-0.8em\TeX}}}
\setcopyright{acmcopyright}
\copyrightyear{2024}
\acmYear{2024}
\acmDOI{XXXXXXX.XXXXXXX}

\acmConference[KDD '24]{Make sure to enter the correct
  conference title from your rights confirmation email}{June 03--05,
  2018}{Woodstock, NY}
\acmISBN{978-1-4503-XXXX-X/18/06}
    
\setcopyright{acmcopyright}
\copyrightyear{2024}
\acmYear{2024}
\acmDOI{XXXXXXX.XXXXXXX}

\acmConference[KDD '24]{The 30th ACM SIGKDD Conference on Knowledge Discovery and Data Mining}{August 25--29,
  2024}{Barcelona, Spain}
\acmBooktitle{KDD '24: The 30th ACM SIGKDD Conference on Knowledge Discovery and Data Mining,
 August 25--29, 2024, Barcelona, Spain} 
\acmPrice{15.00}
\acmISBN{978-1-4503-XXXX-X/18/06}

\usepackage{hyperref}
\usepackage{enumitem}
\usepackage{multirow}
\usepackage{booktabs}
\usepackage{tabularx}
\usepackage{adjustbox}

\begin{document}

\title{AI for Biomedicine in the Era of Large Language Models}

\author{*Zhenyu Bi, *Sajib Acharjee Dip, *Daniel Hajialigol, *Sindhura Kommu, \\ *Hanwen Liu, *Meng Lu, Xuan Wang}
\thanks{*These authors have made equal contributions to this paper. They are listed in alphabetical order by last name.}
\email{{zhenyub, sajibacharjeedip, danielhajialigol, sindhura, liuhwen, menglu, xuanw}@vt.edu}
\affiliation{%
  \institution{Virginia Tech, Blacksburg, VA, USA}
  \country{} 
}

\begin{abstract}
The capabilities of AI for biomedicine span a wide spectrum, from the atomic level, where it solves partial differential equations for quantum systems, to the molecular level, predicting chemical or protein structures, and further extending to societal predictions like infectious disease outbreaks. Recent advancements in large language models, exemplified by models like ChatGPT, have showcased significant prowess in natural language tasks, such as translating languages, constructing chatbots, and answering questions. When we consider biomedical data, we observe a resemblance to natural language in terms of sequences – biomedical literature and health records presented as text, biological sequences or sequencing data arranged in sequences, or sensor data like brain signals as time series. The question arises: Can we harness the potential of recent large language models to drive biomedical knowledge discoveries? In this tutorial, we will explore the application of large language models to three crucial categories of biomedical data: 1) textual data, 2) biological sequences, and 3) brain signals. Furthermore, we will delve into large language models' challenges in biomedical research, including ensuring trustworthiness, achieving personalization, and adapting to multi-modal data representation.
\end{abstract}

\maketitle

\section{Introduction}
The impressive capabilities of AI for biomedicine span a wide spectrum, from the atomic level, where it attempts to solve partial differential equations for quantum systems, to the molecular level, where it accurately predicts the structures of chemicals and proteins, and extends even further, encompassing societal predictions like forecasting infectious disease outbreaks \cite{zhang2023artificial}. Amidst this landscape of possibilities, recent advancements in large language models (LLMs), notably exemplified by models like ChatGPT\footnote{https://chat.openai.com/chat}, have risen to the forefront, demonstrating significant proficiency in tasks tied to natural language. These tasks include language translation, constructing chatbots, and answering questions \cite{yang2023harnessing}.

Interestingly, when we turn our attention to biomedical data, we observe a striking resemblance to natural language in terms of sequences. Biomedical literature and health records are laid out as textual narratives, biological sequences or sequencing data takes the form of molecular or expression sequences, and sensor data like brain signals is inherently sequential time series \cite{wang2021pre, thirunavukarasu2023large}. This observation prompts a compelling question: Can we leverage the advanced LLMs to drive biomedical knowledge discoveries?

In this survey paper, we embark on a journey to explore precisely this intersection—the fusion of cutting-edge large language models with biomedical inquiry. Our exploration zooms in on three pivotal categories of biomedical data: 1) textual data, 2) biological sequences, and 3) brain signals. By drawing inspiration from the transformative capabilities of LLMs, we seek to unravel novel understanding and innovation within each domain.

As we move forward, we further discuss the intricate challenges accompanying AI infusion into biomedical research. The foundation of trustworthiness stands tall—how do we ensure the reliability of AI-enhanced biomedical insights? The concept of personalization emerges as a critical consideration, urging us to tailor LLMs to the specific contours of biomedical investigation. Furthermore, the multi-modal nature of biomedical data motivates us to handle data representations that span across various modalities. 

\section{LLMs on Diverse Biomedical Data}
We comprehensively survey the topics of LLMs for biomedicine based on three pivotal categories of biomedical data: 1) textual data, 2) biological sequences, and 3) brain signals.

\pdfoutput=1
\subsection{LLMs on Biomedical Textual Data}
First, we introduce LLMs in the realm of biomedical textual data, which encompasses diverse domains like biomedical literature \cite{beltagy2019scibert, lee2020biobert, gu2021domain, alrowili2021biom, yasunaga2022linkbert} and electronic health records \cite{alsentzer2019publicly, singhal2022large}. This form of biomedical textual data closely mirrors the fundamental structure of large language models. It finds extensive utility across biomedicine and healthcare, facilitating tasks such as extracting valuable information and responding to queries. The applicability spans a multitude of areas, underpinning biomedical and healthcare information extraction \cite{wang2021chemner, zhong2023reactie, wang2022reactclass} and question-answering \cite{krithara2023bioasq}.

\begin{figure}[t]
    \centering
    \includegraphics[width=0.48\textwidth]{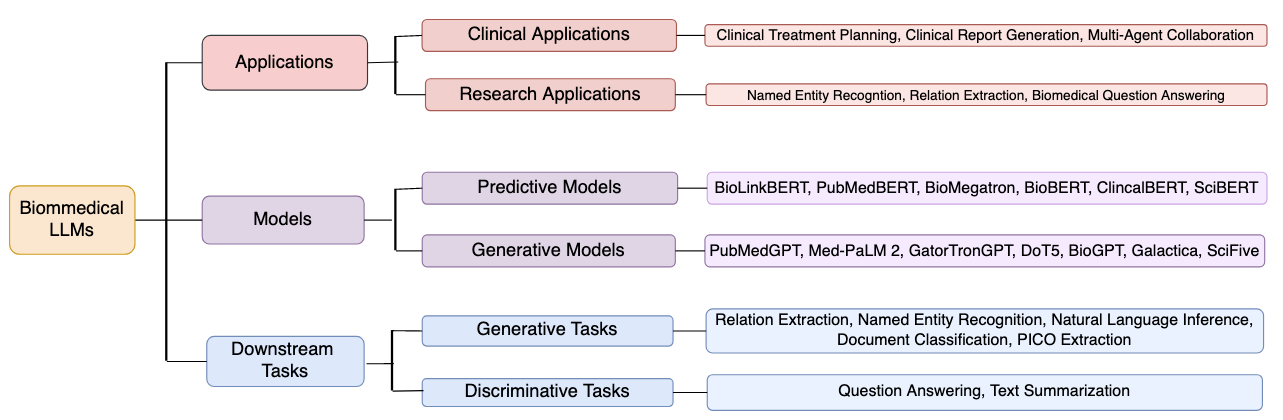}
    \caption{Overview of applications, models, and downstream tasks for biomedical LLMs on textual data.} 
    \vspace{-2mm}
    \label{fig:enter-label}
\end{figure}

\paragraph{SciBERT} SciBERT \cite{beltagy2019scibert} is a BERT-based model \cite{devlin2018bert} pre-trained on scientific publications. Specifically, SciBERT is pre-trained on 1.14 million scientific publications from Semantic Scholar \cite{ammar-etal-2018-construction} via random sampling. SciBERT's vocabulary construction utilizes WordPiece \cite{wu2016google}, the same unsupervised tokenization technique that BERT employs, while also maintaining the same vocabulary size. Only 42\% of the vocabulary across SciBERT and BERT contain the same vocabulary, indicating the difference in the frequency of most common words in the general domain (BERT) and in the scientific domain (SciBERT). SciBERT surpasses BERT in all four evaluated downstream tasks, which include named entity recognition, PICO (Population, Intervention, Comparison, Outcome) extraction, relation extraction, and dependency parsing.  Additionally, SciBERT was state-of-the-art in all four tasks excluding dependency parsing at the time of publication. 

\paragraph{ClinicalBERT} ClinicalBERT \cite{alsentzer2019publicly} is the first framework to pre-train clinical texts using a BERT-based architecture. Approximately 2 million clinical notes, originating from the MIMIC III v1.4 database \cite{johnson2016mimic}, are used for pre-training. The discharge summaries were used for pre-training to generate Discharge BERT, while the clinical note types (nursing, radiology, etc.) were used to train Clinical BERT. Experiments were conducted on natural language inference and named entity recognition tasks. ClinicalBERT demonstrated state-of-the-art results on natural language inference and two of the four named entity recognition datasets at the time of publication. A concern of using ClinicalBERT is the lack of robustness for patient de-identification, as the framework was not able to obtain improvements on the de-identification task.

\paragraph{BioBERT} BioBERT \cite{lee2020biobert} is a BERT-based architecture that is pre-trained on biomedical data. Precisely, BioBERT pre-trains on Pubmed abstracts (4.5 billion words) and PMC publications (13.5 billion words). At the time of publication, experimental results indicate that BioBERT achieves state-of-the-art performance on the majority of biomedical named entity recognition, biomedical question answering, and biomedical relation extraction datasets.

\paragraph{BioMegatron} BioMegatron \cite{shin2020biomegatron} is biomedical LLM based on Megatron-LM \cite{shoeybi2019megatron}, where model parallelism is applied to both self-attention and multi-layer perceptron blocks for each transformer layer. While Megatron-LM is pre-trained on general domain text, BioMegatron is pre-trained on approximately 6.1 billion words, 4.5 billion of which are original from a PubMed abstract dataset, and the remaining 1.6 billion derived from all PMC scientific publications. BioMegatron's size ranges from 345 million to 1.2 billion and also has the flexibility to use vocabularies of either 30,000 or 50,000 subtokens. Their experiments indicate that BioMegatron performs state-of-the-art on all tested named entity recognition and relation extraction benchmarks, mainly with the larger vocabulary.

\paragraph{SciFive} 
SciFive \cite{phan2021scifive} uses a T5 \cite{raffel2020exploring}, a generative text model that is pre-trained millions of PubMed abstraction and PMC scientific publications. Unlike BERT-inspired architectures, SciFive utilizes a Sentence Piece \cite{kudo-richardson-2018-sentencepiece} algorithm for its vocabulary. State-of-the-art results are achieved against baselines such as BioBERT, general-domain T5, and general-domain BERT on biomedical question answering, relation extraction, named entity recognition, and natural language inference.

\paragraph{PubMedBERT} A common assumption within the biomedical LLM literature is that the transfer of general domain knowledge will help domain-specific, in particular biomedical, downstream tasks; however, PubMedBERT \cite{gu2021domain} shows this transfer of knowledge is detrimental to downstream performance: Pre-training using biomedical corpora results in better performance than continual pre-training. This performance improvement is in large part due to PubMedBERT having an in-domain vocabulary, whereas mixed-domain pre-training is subjected to model out of vocabulary words using fragmented subwords. PubMedBERT uses BERT as its backbone and is pre-trained on PubMed articles. Across all 14 datasets stemming from relation extraction, named entity recognition, document classification, question answering, and PICO extraction for evaluation, PubMedBERT achieves state-of-the-art on 12 datasets.

\paragraph{BioLinkBERT} Contrary to the approach of many biomedical LLMs, which typically pre-train using only single documents, BioLinkBERT \cite{yasunaga2022linkbert} utilizes the multiple links between documents, resulting in a multi-task self-supervised pre-training, modeling both document relation extraction and masked language modeling. BioLinkBERT utilizes PubMed, equipped with citations, as its pre-training corpus. To facilitate the document relation extraction task, BioLinkBERT creates a document relation extraction dataset of the pre-training corpus. Each pair of sections (coming from either the same or different documents) falls under one of three classes: random, contiguous, or linked, where "contiguous" indicates the sections (from the same document) are right next to each other, and "linked" indicates two sections (from different documents) are indeed linked. BioLinkBERT achieves state-of-the-art performance on 12 of the 13 different datasets used for evaluations that span 6 different tasks (named entity recognition, PICO extraction, relation extraction, sentence similarity, document classification, and question answering).

\paragraph{Galactica} Galactica \cite{taylor2022galactica} employs a Transformer-based architecture but generates a vocabulary of size 50,000 using byte-pair encoding as well as learnable positional embeddings. The size of Galactica concerning its parameters can range from 125 million to 120 billion. Galactica has multiple tokenization techniques dependent on the modality that's presented to it. This mainly comes in the form of special tokens wrapping around a modality's input. For example, citation information is wrapped around \textsc{[START\_REF]} and \textsc{[END\_REF]} tokens; special tokens \textsc{[START\_AMINO]} and \textsc{[END\_AMINO]} are wrapped around amino acid sequences. Across all types of modalities, Galactica was pre-trained on 106 billion tokens from scientific sources such as scholarly papers, textbooks, and encyclopedias. Galactica either achieves state-of-the-art performance or is within 0.5\% on biomedical question-answering datasets that are evaluated. 

\paragraph{BioGPT} 
BioGPT \cite{luo2022biogpt} is one of the first biomedical GPT-inspired biomedical frameworks. BioGPT is pre-trained on 15 million PubMed articles and employs byte-pair as their vocabulary-building algorithm, resulting in a vocabulary of approximately 42,000 in size. BioGPT is trained with the GPT-2 \cite{radford2019language} backbone due to the sheer size of GPT-3 \cite{brown2020language} (15 billion parameters). As a result, BioGPT has 347 million parameters. Experiments on three biomedical relation extraction datasets and a biomedical question-answer dataset indicate state-of-the-art performance at the time of publication.

\paragraph{DoT5} 
DoT5 \cite{liu2023compositional} proposed a zero-shot domain transfer framework that transfer domain knowledge to the biomedical domain. DoT5 continuously pre-trains using a multi-task setup: combining general-domain task knowledge and biomedical knowledge. Specifically, DoT5's pre-training objective is a linear combination of solving domain-agnostic tasks (natural language inference, summarization, etc.) and masked biomedical language modeling. DoT5 uses T5 \cite{raffel2020exploring} as its backbone and utilizes PubMed abstracts for its in-domain dataset. Results indicate that DoT5 is competitive with state-of-the-art biomedical natural language inference and biomedical summarization tasks.

\paragraph{GatorTronGPT} Considerable stigma associated with applying LLMs like ChatGPT to biomedicine stems from their lack of specialization in the biomedical field. GatorTronGPT \cite{peng2023study} was introduced to alleviate this stigma, as it is pre-trained on billions of clinical texts. Specifically, GatorTronGPT was pre-trained on 277 billion words, 82 billion of which originated from de-identified clinical text \cite{yang2022large} and the remaining 195 billion were obtained from Pile \cite{gao2020pile}, a diverse dataset specifically for language modeling. GatorTronGPT was evaluated on downstream tasks such as biomedical question answering and biomedical relation extraction. To evaluate the quality of the generated texts, billions of synthetic text were generated from GatorTronGPT that were used for fine-tuning a BERT-based model, GatorTronS, on other downstream tasks such as clinical concept extraction, biomedical relation extraction, semantic textual similarity, natural language inference, and question answering. Additionally, it was shown that less than half of 30 synthetic clinical paragraphs generated by GatorTronGPT were identified to be synthetically generated by two physician evaluators at the University of Florida. While GatorTronGPT takes steps to de-identify its clinical data, the authors suggest future studies to validate the robustness of re-identifying patients further.


\paragraph{Med-PaLM 2} Med-PaLM 2 \cite{singhal2023towards} is an extension of Med-PaLM \cite{chung2022scaling} with several improvements. First, Med-PaLM 2 utilizes PaLM 2 \cite{anil2023palm} as the backbone LLM, unlike PaLM which was the backbone of Med-PaLM. Second, to improve reasoning capabilities, Med-PaLM 2 introduces ensemble refinement, a prompting strategy that iteratively feeds in prior generations, consisting of an answer and explanation, as well as the given question and prompt to output a refined answer and generation. Performance on the MedQA medical dataset is state-of-the-art by the time of publication, beating the previous state-of-the-art (Med-PaLM) by over 19\%. Additionally, Med-PaLM 2 performed state-of-the-art or had comparable performance to other clinical datasets, such as PubMedQA, MMLU, and MedMCQA.

\paragraph{\textbf{Applications}:}
There are multiple downstream applications for Biomedical LLMs. We roughly divide them into two categories: clinical applications and research applications.

\paragraph{NLP-related Clinic applications.}
The most widely-researched directions in Biomedical NLP related to clinical are Clinical Treatment Planning, Clinical Reports Generation, and Multi-Agents Collaboration. Clinical Treatment Planning\cite{rajpurkar2022ai} involves diagnosing health issues and determining the suitable treatment course, requiring healthcare professionals to assess objective medical test data and subjective patient symptoms. Incorporating Large Language Models (LLMs) could enhance diagnostic accuracy and broaden access to specialized medical knowledge\cite{thirunavukarasu2023large}. In parallel, Clinical Reports Generation\cite{wu2023generalist}, crucial for documenting patient medical histories, benefits significantly from LLMs\cite{wang2023chatcad}. These models facilitate the creation of reports by summarizing content, referencing relevant information, and generating follow-up diagnostic comments and medication recommendations based on clinical diagnoses\cite{wu2023gpt4vision}. Furthermore, Multi-Agents Collaboration\cite{xi2023rise}, by simulating various diagnostic scenarios, enables a synergy among agents to interpret data and perform complex clinical tasks. This collaboration extends to integrating Electronic Health Records (EHR)\cite{solares2021transfer} and medication history, thus offering real-time management and advisories. Such an approach augments diagnostic accuracy, optimizes the duration of medical procedures, and minimizes patient recovery times\cite{tang2024medagents}, offering a comprehensive and dynamic support system for healthcare providers.

\paragraph{NLP-related Research applications.} 
The most widely-researched directions in Biomedical NLP are information extraction (IE) related tasks, such as Named Entity Recognition (NER) and Relation Extraction (RE), where the model is asked to extract entities (such as genes, drug names, and diseases) or relations of the entities, from text data in biomedical literature. Another prominent direction in the field is Biomedical Question Answering, where the model is asked to answer biomedical-related questions, given heavily professional background text. Since the aforementioned tasks are very domain-specific, fine-tuning or pre-training domain-specific LLMs are most effective when dealing with them \cite{Gao2021APA,Sheikhshab2018IndomainCT,Zhu2018ClinicalCE}. The biomedical domain-specific datasets are often created based on PubMed articles \cite{jin2019pubmedqa,10.1093/bib/bbac282,10.1093/database/baw068}, professional medical examinations \cite{jin2020disease}, or real-life patient health information \cite{johnson2018mimic}, with extensive expert annotation.


\pdfoutput=1
\subsection{LLMs on Biological Sequences}
Next, we extend the application of LLMs to the intricate realm of biological sequence data, where a rich landscape of possibilities emerges. Within this domain, we shift our focus to four major categories of biological sequences: 1) DNA sequences, 2) RNA sequences, 3) protein sequences, and 4) multi-omics sequencing data.

\begin{figure}[t]
    \centering
    \includegraphics[width=0.48\textwidth]{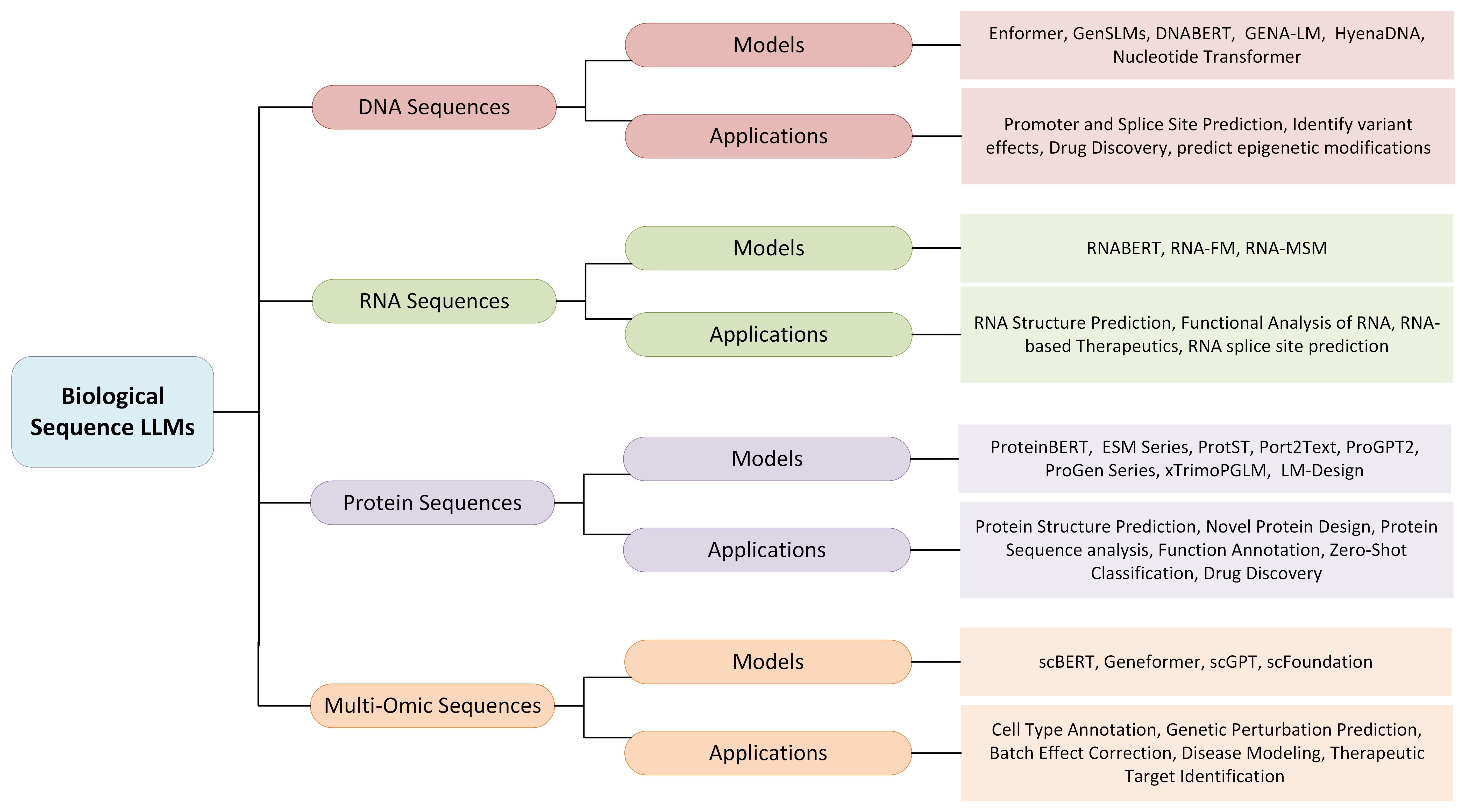}
    \caption{Overview of models and applications for Genomic LLMs on biological sequences.}
    \vspace{-2mm}
    \label{fig:enter-label}
\end{figure}

\subsubsection{\textbf{DNA sequences}:}

The central dogma of molecular biology outlines the flow of genetic information within living organisms, with DNA serving as the source of this crucial information. DNA variants play a significant role in determining heritable traits, influencing aspects related to health and disease. Various innovative works in genomics, such as Enformer \cite{Avsec2021}, Nucleotide Transformer \cite{dalla2023nucleotide}, GenSLMs \cite{zvyagin2022genslms}, DNABERT \cite{ji2021dnabert}, GENA-LM \cite{fishman2023gena}, and HyenaDNA \cite{nguyen2023hyenadna}, are proposed to decode the genetic blueprint of life. 


\paragraph{Enformer} Enformer \cite{Avsec2021} is designed to predict gene expression and chromatin states purely from DNA sequences by integrating long-range interactions. It utilizes transformer modules with self-attention to achieve a larger receptive field, enabling the detection of sequence elements up to 100 kb (kilobases) away, compared with 20 kb for Basenji2 \cite{kelley2020cross} in the genome. By increasing this information flow between distal elements and prioritizing relevant data across the DNA sequence, the model can better capture biological phenomena, such as enhancers that regulate promoters, despite a large DNA-sequence distance between the two. Enformer demonstrates superior performance over the previous model Basenji2 in gene expression prediction across genes and CAGE experiments for protein-coding genes. It excels in predicting tissue and cell-type-specific gene expression, enhancing accuracy in enhancer-promoter prediction and non-coding variant effect prediction. 

\paragraph{GenSLMs} GenSLMs \cite{zvyagin2022genslms}, or Genome-scale Language Models, represent a groundbreaking approach to understanding the evolutionary dynamics of SARS-CoV-2. These models utilize large language models (LLMs) adapted for genomic data to learn the evolutionary landscape of SARS-CoV-2 genomes. By pre-training on a vast number of prokaryotic gene sequences and fine-tuning on SARS-CoV-2-specific data, GenSLMs can accurately and rapidly identify variants of concern. They are among the first whole-genome scale foundation models capable of generalizing to various prediction tasks. The models have been demonstrated to scale efficiently on GPU-based supercomputers and AI-hardware accelerators, achieving remarkable computational performance. This innovative approach aims to enhance public health intervention strategies and inform vaccine development for emerging variants by providing insights into the evolutionary dynamics of SARS-CoV-2.

\paragraph{DNABERT} DNABERT \cite{ji2021dnabert} is a pre-trained BERT model specifically designed for genomic DNA sequences. It addresses the complexity of the gene regulatory code by capturing global and transferable understandings of DNA sequences based on nucleotide contexts. DNABERT excels in data-scarce scenarios, achieving state-of-the-art performance in predicting splice sites and transcription factor binding sites. It allows for direct visualization of nucleotide-level importance and semantic relationships within sequences, enhancing interpretability and identification of genetic variants. 

\paragraph{GENA-LM} GENA-LM \cite{fishman2023gena} represents a significant advancement in the field of computational genomics, introducing a suite of open-source, transformer-based foundational DNA language models specifically designed to handle long DNA sequences. These models are capable of processing input lengths up to 36,000 base pairs, a substantial increase over previous architectures, which struggled with the extensive contextual information spread across such long sequences. Key features of GENA-LM include its use of Byte Pair Encoding (BPE) tokenization, which differs from the k-mer approach used in earlier models like DNABERT. This allows for a more efficient representation of DNA sequences, facilitating the processing of longer inputs. The models have been pre-trained on the latest T2T human genome assembly \cite{Nurk2022-nq}, incorporating a wide range of genomic data to enhance their predictive capabilities.

\paragraph{HyenaDNA} HyenaDNA \cite{nguyen2023hyenadna} is a groundbreaking genomic foundation model designed for long-range sequence modeling at single nucleotide resolution. It addresses the limitations of previous models by enabling the processing of ultralong sequences directly at the single nucleotide level without the need for tokenizers. HyenaDNA is significantly smaller than previous genomic models, yet it demonstrates exceptional performance, surpassing state-of-the-art benchmarks on various datasets with fewer parameters and pre-training data. This model scales sub-quadratically in sequence length, allowing for efficient training and processing of long DNA sequences. Additionally, HyenaDNA shows promise in species classification tasks and offers exciting prospects for synthetic regulatory element design, gene identification, and protein complex modeling. Despite its early-stage limitations related to training on a single human genome and focusing solely on DNA, HyenaDNA's potential for personalized genomic analysis and innovative applications in drug discovery is immense.

\paragraph{Nucleotide transformer} The Nucleotide Transformer model \cite{Avsec2021} is a robust foundational model pre-trained on raw DNA sequences from diverse human genomes and various species. This model tokenizes sequences into six-character words (k-mers of length 6) and utilizes the BERT methodology for training. It is then applied to 18 downstream tasks covering various predictions like promoter prediction, splice site donor and acceptor prediction, histone modifications, and more. Predictions are made through probing or light fine-tuning, providing accurate molecular phenotype predictions even in low-data settings by generating context-specific representations of nucleotide sequences. The Nucleotide Transformer outperforms specialized methods on multiple tasks, showcasing a focus on key genomic elements such as gene expression regulators. Models trained on genomes from different species show superior performance compared to those trained solely on human sequences in various human prediction tasks, indicating that diverse species training enhances generalization in human-based predictions. 

\paragraph{\textbf{Applications:}}
Genomic foundation models play a crucial role in various tasks, including predicting gene expression, deciphering DNA sequences for biological functions, identifying genomic region structures, and predicting genetic elements such as promoter regions, enhancers, and transcription factor binding sites. These models are instrumental in understanding transcriptional regulation and its connection to non-coding genetic variants linked to human diseases and traits. For example, Enformer can directly predict enhancer-promoter interactions from DNA sequences \cite{Avsec2021}, competing with methods that rely on experimental data. Additionally, GenSLMs, trained on over 110 million prokaryotic gene sequences and fine-tuned for SARS-CoV-2, can accurately and rapidly identify variants of concern \cite{zvyagin2022genslms}, which also represents the generalization ability of these models to different prediction tasks. DNABERT, fine-tuned for specific tasks like predicting promoters, transcription factor binding sites, and splice sites, has been benchmarked against state-of-the-art tools in the field \cite{ji2021dnabert}. These genomic models that can accurately identify promoter regions and splice sites, aid in understanding gene regulation and RNA transcript processing. They can also predict epigenetic modifications \cite{Dunham2012} and the effects of DNA variations on genes and protein function \cite{doi:10.1073/pnas.2311219120}, helping identify genetic risk factors. Moreover, DLM models contribute to understanding how non-coding DNA influences gene expression \cite{Zhou2015-pp} and disease \cite{10.1093/nar/gkz972} in different cell types, which is crucial for human genetics and disease studies.

\subsubsection{\textbf{RNA sequences}:} 

Ribonucleic acid (RNA) is another critical component of molecular biology, playing various essential roles in fundamental biological processes. In recent years, RNA has emerged as an intriguing target for drug development \cite{Garner2023-oj}, underscoring the importance of enhancing our understanding of its structures and functions. We embrace the innovative contributions highlighted in foundation models such as RNABERT [\cite{10.1093/nargab/lqac012}], RNA-FM \cite{chen2022interpretable}, and RNA-MSM \cite{zhang2023multiple}, which are pivotal in elucidating the intricate mechanisms governing biological processes mediated by RNA. These advancements empower us to decipher the complex orchestration of biological processes by RNA molecules.

\paragraph{RNABERT} RNABERT \cite{10.1093/nargab/lqac012} proposes an effective method for embedding RNA bases by applying the pre-training approach of BERT to non-coding RNA (ncRNA). The research aims to develop a robust RNA base embedding technique for tasks like RNA structural alignment and clustering using deep representation learning. By leveraging pre-training algorithms, the researchers aim to create semantically rich representations of RNA bases to improve the accuracy of structural alignment and clustering tasks. This method of informative base embedding incorporates contextual information and secondary structure details of RNA sequences, resulting in superior accuracies compared to existing methods such as GraphClust \cite{heyne2012graphclust}, EnsembleClust \cite{saito2011fast}, and CNNclust \cite{aoki2018convolutional}. The approach combines informative base embedding with a simple Needleman–Wunsch alignment algorithm to calculate structural alignments with improved time complexity, making it a significant advancement in RNA sequence analysis.

\paragraph{RNA-FM} RNA-FM \cite{chen2022interpretable} is a novel interpretable computational model proposed to predict RNA structure and function accurately using unannotated data. It leverages self-supervised learning on 23 million non-coding RNA sequences to infer sequential and evolutionary information without labels. It has shown effectiveness in predicting secondary/3D structures, SARS-CoV-2 genome structure, protein-RNA binding preferences, and gene expression regulation. Despite being trained on unlabelled data, RNA-FM significantly improves RNA structural and functional modeling results, making it a foundational model for the field.

\paragraph{RNA-MSM} RNA-MSM \cite{zhang2023multiple} is an RNA language model that leverages homologous RNA sequences to interpret and predict RNA structures more effectively than previous methods. It is distinguished by its use of unsupervised learning on multiple sequence alignments (MSAs) derived from RNAcmap, a tool that outperforms traditional methods like Rfam in identifying homologous sequences due to its comprehensive and automatic pipeline. This approach allows RNA-MSM to capture evolutionary information and structural nuances of RNA sequences, which are less conserved and thus more challenging to analyze compared to proteins. The model generates two-dimensional attention maps and one-dimensional embeddings that contain rich structural information. These outputs can be directly correlated with RNA structural features, such as 2D base pairing probabilities and 1D solvent accessibilities, with high accuracy. The study demonstrates that RNA-MSM significantly outperforms existing techniques, including SPOT-RNA2 and RNAsnap2, in predicting these structural features after fine-tuning. This research underscores the potential of RNA-MSM to revolutionize RNA structure prediction and functional analysis by incorporating evolutionary and structural insights from homologous sequences. 

\paragraph{\textbf{Applications:}}

RNA language models have a wide range of applications in RNA research and biotechnology. These models utilize evolutionary information from homologous sequences to accurately interpret RNA sequences, making them valuable in tasks such as RNA structure \cite{yamada2022prediction}, function \cite{zhang2023multiple}, and RNA-protein interaction prediction \cite{chen2022interpretable}. They can predict crucial aspects of RNA structures, such as 2D base pairing probabilities and 1D solvent accessibilities \cite{zhang2023multiple}, which are essential for understanding RNA function and cellular interactions. Additionally, RNA language models facilitate the design of RNA-based therapeutics \cite{Li2023.09.09.556981}. Understanding RNA structures helps identify novel drug targets and design RNA-based drugs, speeding up drug discovery processes. These pre-trained models can be fine-tuned for various downstream tasks related to RNA structure and function, including RNA 2D/3D structure prediction \cite{zhang2023multiple}, RNA structural alignment and family clustering \cite{10.1093/nargab/lqac012}, and RNA splice site prediction \cite{Chen2023.01.31.526427} from RNA sequences.

\subsubsection{\textbf{Protein sequences}:}

Venturing into the complex realm of proteins, we embrace luminous works such as \cite{rives2021biological}, \cite{bepler2021learning}, \cite{brandes2022proteinbert}, \cite{madani2023large}, \cite{lin2023evolutionary}, \cite{zheng2023structure}, and \cite{xu2023protst}. These endeavors illuminate the path to unraveling the intricate choreography of molecular functions and interactions. The protein LLMs have a wide application in functional protein generation \cite{leinonen2004uniprot} and protein structure prediction \cite{suzek2015uniref}.

Protein language models can be broadly classified into three main categories based on their functionalities: 1) structure and function understanding: interpreting protein sequences to reveal their structures and functions through evolutionary data or textual enhancements, 2) protein sequence generation: creating novel protein sequences using generative modeling for unexplored spaces or precise design, and 3) multi-modal integration for enhanced understanding and generation: combining various data types to generate protein functions or achieve a unified understanding and generation of protein sequences.

\paragraph{Structure and Function Understanding models} These models are primarily concerned with interpreting existing protein sequences to deduce their structures, functions, or both, using vast datasets of known proteins. ProteinBERT \cite{brandes2022proteinbert} is a deep-learning model designed for understanding protein functions and sequences. It combines language modeling with Gene Ontology (GO) \cite{ashburner2000gene} annotation predictions in a self-supervised training approach. Trained on over 106 million proteins from the UniProtKB/UniRef90 database \cite{boutet2007uniprotkb, suzek2007uniref}, it efficiently captures detailed and broad protein features using a transformer-like architecture. 
ESM models are noted for their ability to capture deep evolutionary signals and predict protein structures with high accuracy, contributing significantly to structural biology. ESM-1b \cite{rives2021biological}, utilizing a Transformer architecture, harnesses unsupervised learning from 250 million protein sequences to predict biological properties. It offers a versatile and comprehensive approach to protein analysis, with applications ranging from structure prediction to functional annotation, by capturing deep evolutionary signals across various biological research tasks. ESM-2 \cite{lin2023evolutionary} significantly advances over ESM-1b by utilizing a scaled-up model with 15 billion parameters (compared to 650 million) and a richer UniRef90 dataset to predict atomic-level protein structures directly from sequences. This enhancement not only allows for high-resolution structure predictions and the construction of the ESM Metagenomic Atlas but also achieves a notable speed improvement and wider applicability in structural genomics and protein design, thus enhancing diversity and accuracy in structural predictions. 
ProtST \cite{xu2023protst} enhances protein language models by merging protein sequences with biomedical texts from the ProtDescribe dataset, sourced from Swiss-Prot \cite{bairoch2000swiss}, to provide a comprehensive understanding of protein functions and properties. This multi-modality learning not only addresses gaps in sequence-only models but also enhances protein representations for a variety of applications, including function annotation and localization prediction. xTrimoPGLM \cite{chen2024xtrimopglm} revolutionizes protein sequence analysis and generation by merging autoencoding and autoregressive objectives within a massive 100 billion parameter framework, trained on about 940 million unique protein sequences. 

\paragraph{Sequence Generation based models}
In the quest to generate novel protein sequences with the potential to fold into functional proteins, several key players have emerged, notably ProGen \cite{madani2023large}, ProGen2 \cite{nijkamp2023progen2}, ProtGPT2 \cite{ferruz2022protgpt2}, and LM-Design \cite{zheng2023structure}. ProGen starts the journey by training on a vast array of 280 million protein sequences, using special tags to make proteins with specific functions \cite{madani2023large}. ProGen2 takes this further by using a much larger model to capture more complex protein patterns and making it possible to predict protein fitness without extra steps \cite{nijkamp2023progen2}. ProtGPT2, inspired by language models, generates proteins that could exist in nature, trained on 50 million sequences for wide exploration. While ProGen series focus on making diverse and functional proteins, ProtGPT2 pushes the limits of designing completely new proteins, showing the power of deep learning in understanding and creating life's building blocks \cite{ferruz2022protgpt2}. LM-Design innovates in protein design by infusing language models with structural adapters, enhancing the generation of sequences. This addresses both the creativity seen in ProtGPT2 and the functional specificity of the ProGen series with an added focus on structural viability \cite{zheng2023structure}.

\paragraph{Multi-modal Integration models}

ProtST \cite{xu2023protst}, Prot2Text \cite{abdine2023prot2text}, and xTrimoPGLM \cite{chen2024xtrimopglm} exemplify the fusion of sequence, structure, and textual data to redefine protein function understanding and design. ProtST integrates dual language models, specifically the ESM series and ProtBert for protein sequences, alongside PubmedBERT \cite{gu2021domain} for biomedical texts, to synergize textual descriptions with sequence data \cite{xu2023protst}. Prot2Text utilizes Graph Neural Networks and Transformers to generate descriptive texts of protein functions, integrating structural and sequence data with textual annotations. This approach not only enhances function annotation and drug discovery efforts but also enriches databases with contextually rich protein descriptions, setting a new standard for how protein functions are predicted and understood \cite{abdine2023prot2text}. xTrimoPGLM, while its primary category is understanding and generation, its methodological approach of combining masked language modeling (MLM) and general language modeling (GLM) objectives allows it to handle multimodal data for enhanced protein sequence analysis \cite{chen2024xtrimopglm}.

\paragraph{\textbf{Applications}:}
In the rapidly evolving field of computational biology, all these protein sequence based large language methods \cite{rives2021biological, lin2023evolutionary, abdine2023prot2text, brandes2022proteinbert,xu2023protst,chen2024xtrimopglm} are pushing the boundaries of how we understand, design, and utilize proteins. The application of these models spans from detailed protein function annotation \cite{brandes2022proteinbert} and structure prediction \cite{rives2021biological, lin2023evolutionary} to the high-throughput design of novel proteins with specific functionalities \cite{ferruz2022protgpt2, madani2023large, nijkamp2023progen2}, significantly impacting drug discovery, synthetic biology, and therapeutic development \cite{chen2024xtrimopglm, lin2023evolutionary}. Looking forward, the integration of even more diverse datasets, the refinement of models to enhance accuracy and efficiency, and the exploration of uncharted areas of the protein universe remain pivotal directions. 

\subsubsection{\textbf{Multi-omics sequencing data}:}

Within these domains, the transformative capabilities of LLMs manifest in high-impact downstream applications. From predicting molecular structures to forecasting molecule interactions, and from unraveling molecule functions to drawing poignant associations with disease progression processes, LLMs guides us towards a deeper comprehension of life's building blocks. In multi-omic sequencing, models such as scBERT\cite{yang2022scbert}, Geneformer\cite{theodoris2023transfer}, scFoundation\cite{hao2023large} and scGPT\cite{cui2024scgpt} use advanced pre-trained language models to deeply analyze cellular biology and genetics, showcasing the fusion of cutting-edge computation with biological understanding.

\paragraph{scBERT}
scBERT \cite{yang2022scbert}, a transformer-based deep neural network model, leverages extensive unlabeled single-cell RNA-seq data for pre-training, addressing the challenges of cell type annotation by capturing latent gene-gene interactions and overcoming limitations such as batch effects and the absence of curated marker genes. Through self-supervised learning and supervised fine-tuning, scBERT significantly enhances cell type annotation accuracy, robustness against batch effects, and interpretability over existing methods. Its versatility extends to discovering novel cell types and understanding complex gene expression patterns, offering a comprehensive tool for single-cell omics analysis.

\paragraph{Geneformer}
Geneformer \cite{theodoris2023transfer}, pre-trained on about 30 million single-cell transcriptomes, employs a context-aware, attention-based framework for versatile applications in network biology, including disease modeling and therapeutic target identification, especially in data-scarce areas. It differentiates itself with a self-supervised learning strategy for fundamental insights into network dynamics. This emphasizes its utility in a broader range of downstream tasks such as chromatin dynamics predictions. In comparison, scBERT, focusing on cell type annotation, demonstrates its strength in handling batch effects and discovering novel cell types through its transformer architecture. Both models offer significant advancements over traditional methods, with Geneformer offering a broader application scope and scBERT excelling in cell type specificity and interpretability.

\paragraph{scFoundation}
scFoundation \cite{hao2023large} redefines the landscape of single-cell analysis with its large-scale pre-trained model, trained on an unprecedented dataset of over 50 million human single-cell transcriptomes. Through its unique asymmetric architecture and the introduction of a read-depth-aware pre-training task, the model efficiently addresses the complexities of single-cell RNA-seq data, including its sparsity and high dimensionality. Distinguished by its vast scale of trainable parameters and extensive data coverage, scFoundation delivers superior performance across a spectrum of downstream applications, from enhancing gene expression to predicting drug responses. 

\paragraph{scGPT}
scGPT \cite{cui2024scgpt}, trained on 33 million cells, significantly advances cell biology and genetics research with its large-scale data analysis. It differentiates itself by applying generative pre-training on extensive multi-omics data for a comprehensive analysis. Using a self-attention transformer, scGPT excels in analyzing complex cell data, enabling it to perform exceptionally well in identifying cell types, predicting genetic changes, and integrating multi-omics data. Its ability to derive insights into gene-gene interactions and demonstrate a scaling effect of improved performance with larger datasets highlights its potential for single-cell omics research.

\paragraph{\textbf{Applications}:}
The integration of advanced computational models like scBERT \cite{yang2022scbert}, scFoundation \cite{hao2023large}, scGPT \cite{cui2024scgpt}, and Geneformer \cite{theodoris2023transfer} into single-cell and multi-omics analysis heralds a new era in precision biology. These models, each with its unique approach ranging from cell type annotation\cite{yang2022scbert, cui2024scgpt} to the elucidation of complex gene networks, offer comprehensive tools for dissecting the intricacies of cellular functions. Future research directions anticipate the convergence of these models into a cohesive analytical framework, enhancing our capability to predict disease mechanisms and responses to treatment based on sophisticated cell-level predictions \cite{hao2023large, theodoris2023transfer}. As the pool of omics data expands, these models will continue to evolve, driving forward the discovery of novel biological mechanisms and therapeutic targets \cite{theodoris2023transfer}.

\pdfoutput=1
\subsection{LLMs on Brain Signals}

Last, we delve into the fascinating realm of applying LLMs to brain signals. In this section, we focus on two topics: brain LLM building and downstream applications in brain-to-text translation.
We start with introducing various pre-training techniques for brain EEG signal representation learning, including BrainBERT (self-supervised representation learning for intracranial recordings) \cite{wang2022brainbert}, MMM (topology-agnostic EEG representation learning) \cite{yi2023learning}, and LaBraM (generic representation learning for tremendous EEG data from various sources) \cite{jiang2024large}. 

\begin{figure}[t]
    \centering
    \includegraphics[width=0.48\textwidth]{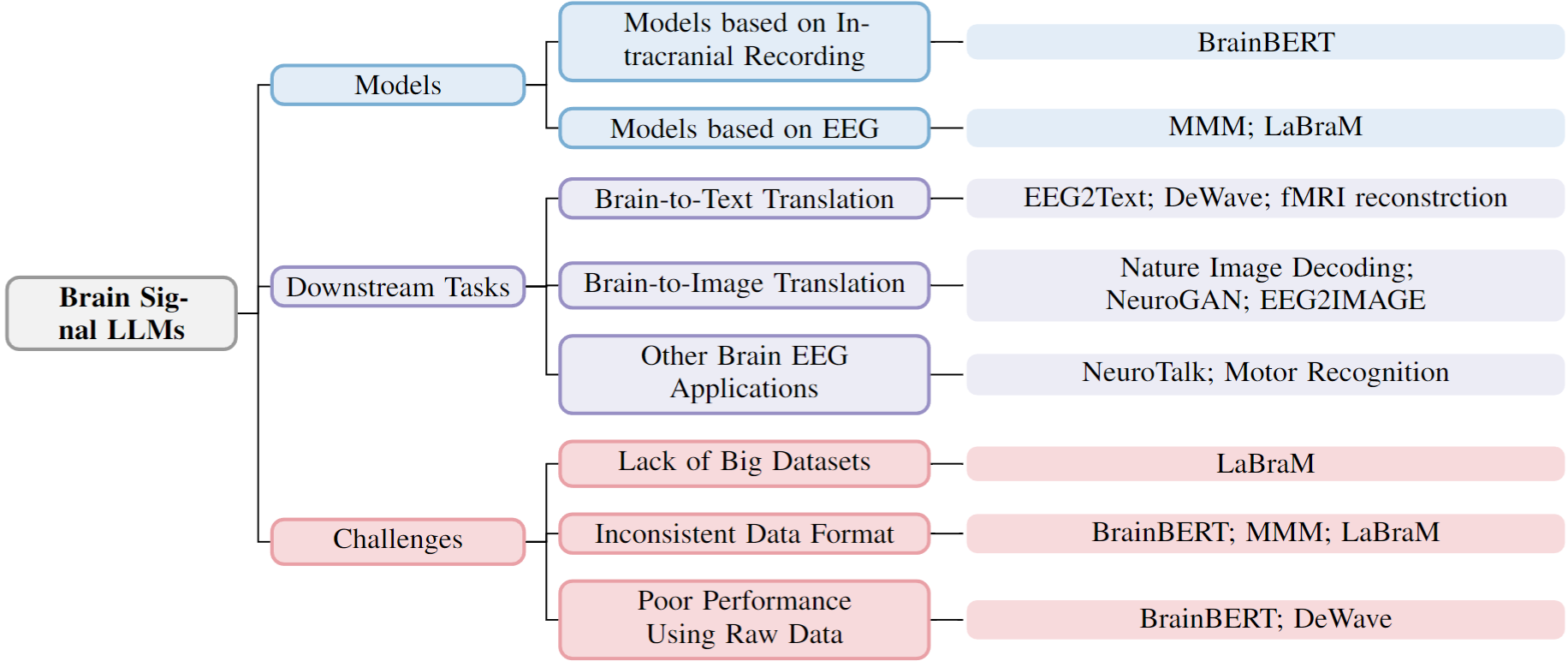}
    \caption{Overview of models, downstream tasks and challenges for brain signal LLMs.}
    \vspace{-2mm}
    \label{fig:brain signal llm}
\end{figure}

\paragraph{BrainBERT} BrainBERT \cite{wang2022brainbert} is a Transformer-based model designed for decoding intracranial field potential recordings. Pretrained unsupervised on large datasets of neural activities recorded while subjects watched movies, BrainBERT facilitates significant improvements in decoding neural signals, addressing the challenges posed by the scarcity and variability of neural data. It excels in generalizing across different subjects and electrode locations, a notable advantage over traditional methods. Employing techniques like super-resolution spectrograms and contextual representation, BrainBERT outperforms existing models, offering enhanced accuracy with less data. BrainBERT demonstrates superior performance across various neural decoding tasks, establishing a new benchmark in the field compared to standard linear decoders and deeper neural networks. The model not only improves brain-computer interface technologies but also enables novel analyses, such as exploring the intrinsic dimensionality of brain computations.

\paragraph{MMM} MMM \cite{yi2023learning} is a novel framework for EEG pre-training. MMM achieves this by mapping different EEG channel selections onto a unified topology and employing strategies such as Multi-dimensional Positional Encoding, Multi-level Channel Hierarchy, and Multi-stage Pre-training. This approach facilitates topology-agnostic EEG representation learning, enhancing cross-dataset generalizability. Evaluations on SEED \cite{seed} and SEED-IV \cite{seediv} datasets demonstrate MMM's superior performance in emotion recognition tasks, outperforming traditional and deep learning baselines. Specifically, MMM surpasses previous state-of-the-art methods in subject-dependent classification, showcasing its robust transferability across different EEG datasets and sensor configurations. The results validate MMM's effectiveness in leveraging larger datasets for extracting more generalized representations.

\paragraph{LaBraM}  LaBraM \cite{jiang2024large} is a novel Large Brain Model, which marks a significant advancement in EEG-based deep learning through its pre-training on over 2500 hours of diverse EEG data. LaBraM overcomes the limitations of EEG data variability and volume by transforming EEG signals into unified channel patches and employing vector-quantized neural spectrum prediction for efficient learning. The model's architecture enables the effective learning of EEG signal representations, addressing both temporal and spatial features. LaBraM is evaluated on various downstream tasks such as abnormal detection, event type classification, emotion recognition, and gait prediction. In these evaluations, LaBraM consistently outperforms state-of-the-art methods across multiple metrics, underscoring its superior ability to generalize from large-scale EEG data. Moreover, the model exhibits scalability with increasing dataset sizes, suggesting potential for further performance improvements with even larger datasets. This research not only sets new benchmarks for EEG-based analyses but also opens up new avenues for deep learning applications in neuroscience and medical diagnostics.

\paragraph{\textbf{Applications}:} We further discuss a broad range of brain LLM applications in brain-computer interfaces (BCIs). 

\paragraph{Brain-to-Text Translation:} 

Previous research on converting brain signals to text, as described in papers\cite{herff2015brain,sun2019towards,anumanchipalli2019speech,makin2020machine,panachakel2021decoding,moses2021neuroprosthesis,nieto2022thinking}, has shown notable success with small and restricted vocabularies. However, these studies have struggled to achieve similar accuracy levels with larger and unrestricted vocabularies.  Building upon this foundation, we further introduce an exciting topic of open-vocabulary brain-to-text translation \cite{wang2022open}, including recent work DeWave (discrete encoding for EEG-to-text translation) \cite{duan2023dewave} and continuous language reconstruction from fMRI images \cite{tang2023semantic}. Starting with EEG2TEXT \cite{wang2022open}, this paper presents a method of open vocabulary EEG-To-Text decoding and zero-shot sentence sentiment classification . Tested on the ZuCo dataset \cite{zuco}, the author utilizes pre-trained language models like BART \cite{bart}, achieving significant advancements with a 40.1\% BLEU-1 score for EEG-To-Text decoding and a 55.6\% F1 score for zero-shot EEG-based ternary sentiment classification, which notably surpass the supervised baselines. Further advancing the field, DeWave \cite{duan2023dewave} is a novel framework for converting EEG signals into text by leveraging a discrete codex encoding. DeWave utilizes a quantized variational encoder to transform EEG waves into discrete representations, aligning them with pre-trained language models for enhanced translation accuracy. The result surpassing previous baselines including EEG2TEXT. In a parallel vein, the paper \cite{tang2023semantic} explores a non-invasive brain-computer interface decoding continuous natural language from fMRI brain recordings.  Employing a unique approach, the model addresses fMRI's low temporal resolution by generating candidate word sequences, scoring them against brain responses to identify the most likely stimuli being heard or imagined. 

\paragraph{Brain-to-Image Translation:} Image Translation is also a mainstream Downstream Tasks of brain signals. Some recent interesting work includes nature image decoding \cite{song2024decoding} NeuroGAN \cite{NeuroGAN} and EEG2IMAGE \cite{EEG2IMAGE}. Specifically, nature image decoding \cite{song2024decoding} introduces a self-supervised framework, which leverages a large and diverse EEG-image dataset \cite{imageeeg}. This dataset is used in conjunction with a novel approach that applies contrastive learning to align features extracted from image stimuli and corresponding EEG responses, significantly advancing the field of non-invasive brain-computer interfaces. The framework employs self-attention and graph attention modules within the EEG encoder to enhance spatial feature extraction, reflecting the spatial dynamics of brain activity related to object recognition. NeuroGAN \cite{NeuroGAN} introduces a sophisticated method for transforming EEG signals into visual images, leveraging a specialized architecture to enhance image synthesis from brain activity data. Central to NeuroGAN is a cross-modality encoder-decoder structure, which effectively compresses EEG features into a latent space and reconstructs corresponding images, focusing on capturing the complex relationship between neural signals and visual stimuli.
The method also incorporates a perceptual loss function, utilizing a pre-trained image classifier to measure the perceptual similarity between generated and real images. Similarly, EEG2IMAGE \cite{EEG2IMAGE} introduces an innovative framework for synthesizing images from EEG signals. The framework uses small EEG datasets to learn features via a contrastive learning approach and synthesizes images using a modified conditional Generative Adversarial Network. Specifically, the framework employs semi-hard triplet loss for feature extraction from EEG signals, ensuring that signals from similar images are closer to the learned feature space, leading to more accurate image reconstruction.

\paragraph{Other Brain EEG Applications:} In addition to text translation and image translation, brain signal decoding also performs well in other downstream tasks. In the field of voice decoding, NeuroTalk \cite{voicere} presents a novel model, for transforming non-invasive EEG signals of imagined speech into audible voice outputs. The model combines multi-receptive residual modules with recurrent neural networks to process brain signals effectively. Notably, the model could generate words not included in the training dataset, indicating its potential for broader vocabulary coverage. In the motion decoding domain, Motor Recognition \cite{motoreeg} explores using EEG data for recognizing motor activities. The research employs an innovative ensemble approach combining stacked bidirectional long short-term memory (BiLSTM) with long short-term memory (LSTM) networks alongside a newly proposed EEG-transformer network to classify 17 different everyday motor activities.  Their ensemble achieves a classification accuracy of 98.5\%, which has a substantial advancement over existing state-of-the-art methodologies.

\section{Conclusions and Future Directions} 
As a conclusion, this is a comprehensive survey of large language models for biomedicine, focusing on three pivotal data types: 1) textual data, 2) biological sequences, and 3) brain signals. There are also significant new challenges that come with using AI in biomedical research. One big challenge is making sure that the biomedical insights enhanced by AI are reliable and trustworthy, including model explainability and interpretability, model robustness to adversarial attacks, model bias towards different populations, and data privacy issues. Another challenge is the personalization of LLMs, which means adjusting LLMs to fit the specific needs of different personalized data. For example, there is a large individual variance in brain signals when different people are thinking of the same word under the same context. Instead of using one LLM to fit everyone, can we construct personalized LLMs based on different brain patterns for different people? The last challenge is the multi-modality data. Since biomedical information can be very varied, we learn how to handle different types of data in a skillful and effective way. For example, Google has announced Med-PaLM-2 \cite{singhal2023towards} that integrates image, text, and genome data in the electronic health record, declaring an expert-level ability for medical question answering. Can we develop more effective and efficient methods to integrate multi-modal and multi-omic LLMs into one powerful unified LLM? This survey paper serves as a foundation step to bridge the gap and encourages both theoretical researchers and practitioners in LLMs to look into real-world scientific applications.

\clearpage
\section*{Acknowledgement}
This work is sponsored by the Commonwealth Cyber Initiative, Children’s National Hospital, Fralin Biomedical Research Institute (Virginia Tech), Sanghani Center for AI and Data Analytics (Virginia Tech), Virginia Tech Innovation Campus, and a generous gift from the Amazon + Virginia Tech Center for Efficient and Robust Machine Learning.

\bibliographystyle{plain}
\bibliography{ref}

\end{document}